\documentclass[conference]{IEEEtran}
\IEEEoverridecommandlockouts
\usepackage{cite}
\usepackage[utf8]{inputenc}
\usepackage{amsmath,amssymb,amsfonts}
\usepackage{algorithmic}
\usepackage{graphicx}
\usepackage{textcomp}
\usepackage{xcolor}
\usepackage{savesym}
\savesymbol{checkmark}
\usepackage{lipsum}
\usepackage[ruled,vlined]{algorithm2e}
\usepackage{enumerate}
\usepackage{tabularx,booktabs}
\usepackage{dingbat}
\usepackage{diagbox}
\usepackage[hidelinks]{hyperref}
\usepackage{multicol}
\usepackage{float}

\restylefloat{table}
\renewcommand{\figurename}{Figure}
\usepackage{fancyhdr} 
\def\BibTeX{{\rm B\kern-.05em{\sc i\kern-.025em b}\kern-.08em
    T\kern-.1667em\lower.7ex\hbox{E}\kern-.125emX}}

 \newcommand{\daffy}[0]{\includegraphics[width=.04\textwidth]{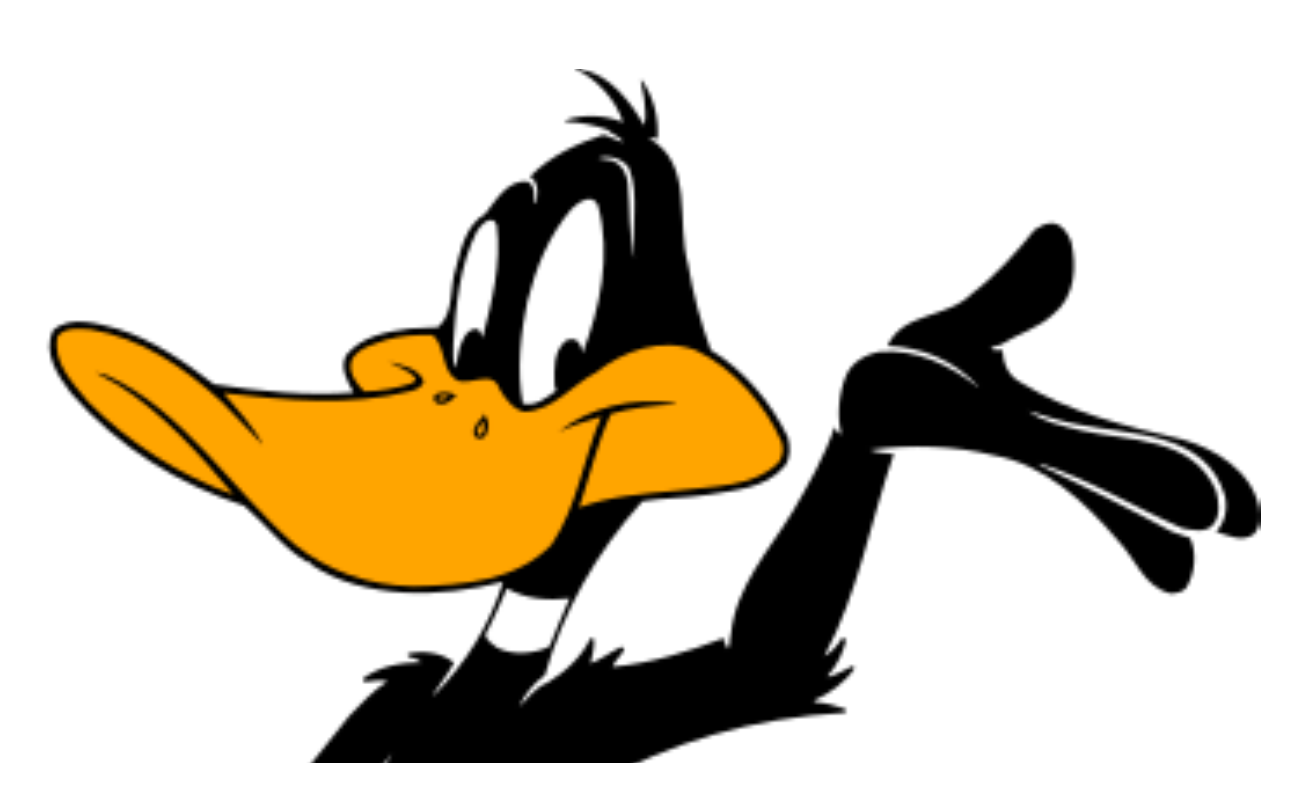}}   
\begin{document}
\pagestyle{fancy}

\title{A Comparative Study of Transformer-Based Language Models on Extractive Question Answering}

\author{%
  Kate Pearce\IEEEauthorrefmark{1},
  Tiffany Zhan\IEEEauthorrefmark{1},

  Aneesh Komanduri\IEEEauthorrefmark{2},
  Justin Zhan\IEEEauthorrefmark{2}
  \\
  \IEEEauthorblockA{%
    \IEEEauthorrefmark{2}Department of Computer Science and Computer Engineering, University of Arkansas
  }
  \IEEEauthorblockA{
    \IEEEauthorrefmark{1} Army Educational Outreach Program UNITE
  }
}

\maketitle
\begin{abstract}
Question Answering (QA) is a task in natural language processing that has seen considerable growth after the advent of transformers. There has been a surge in QA datasets that have been proposed to challenge natural language processing models to improve human and existing model performance. Many pre-trained language models have proven to be incredibly effective at the task of extractive question answering. However, generalizability remains as a challenge for the majority of these models. That is, some datasets require models to reason more than others. In this paper, we train various pre-trained language models and fine-tune them on multiple question answering datasets of varying levels of difficulty to determine which of the models are capable of generalizing the most comprehensively across different datasets. Further, we propose a new architecture, BERT-BiLSTM, and compare it with other language models to determine if adding more bidirectionality can improve model performance. Using the F1-score as our metric, we find that the RoBERTa and BART pre-trained models perform the best across all datasets and that our BERT-BiLSTM model outperforms the baseline BERT model.
\end{abstract}


\section{Introduction}

\IEEEPARstart{E}xtractive Question Answering is the task of extracting a span of text from a given context paragraph as the answer to a specified question. Question Answering is a task in natural language processing (NLP) that has seen considerable progress in recent years with applications in search engines, such as Google Search and chatbots, such as IBM Watson. This is due to the proposal of large pre-trained language models, such as BERT \cite{devlin2019bert},  which utilize the Transformer \cite{vaswani2017attention} architecture to develop robust language models for a variety of NLP tasks specified by benchmarks, such as GLUE \cite{wang2019glue} or decaNLP \cite{mccann2018natural}. Additionally, new datasets, including SQuAD \cite{rajpurkar2018know} have introduced more complex questions with inference based context to the question answering task. Recent work has shown to be productive in tackling the task of question answering, but the task is nowhere near solved. With the introduction of datasets, such as QuAC \cite{choi2018quac} and NewsQA \cite{trischler2017newsqa} that rely much more on reasoning, it becomes challenging to generalize previous well performing QA models to different datasets. 

With more accessibility to computational power, several variants of BERT have been proposed for different domains. For example, BERTweet \cite{nguyen2020bertweet} is a bidirectional transformer model trained on twitter data and can be particularly useful for analyzing social media data. CT-BERT \cite{muller2020covidtwitterbert} is another domain-specific variant that was specifically pre-trained on Covid-19 tweets. Before transformers changed the landscape of NLP, sequence-to-sequence models such as the Recurrent Neural Network and Long-Short Term Memory (LSTM) \cite{HochSchm97} were the state-of-the-art, beating out nearly all other trivial approaches. For the task of question answering, one of the first sequence models proposed was MatchLSTM \cite{wang2016machine}. 

In this paper, we study the difference between various pre-trained transformer-based language models to analyze how well they are able to generalize to datasets of varying levels of complexity when fine-tuned on the question answering task. Furthermore, we propose an ensemble model architecture using BERT and BiLSTM and evaluate its performance against standard pre-trained models on extractive question answering. 

The paper is organized as follows. Section II discusses background pertaining to question answering and pre-trained language models. We discuss our approach and propose our model architecture in Section III. Section IV contains our experimental procedure including datasets and choice of hyperparameters. Section V will be an analysis of our experimental results. Finally, we conclude the paper and discuss future work in section VI.

\section{Background}
\label{sec:background}

\subsection{Transformers}
Recurrent neural networks and gated recurrent neural networks have been established as effective approaches to natural language processing tasks like machine translation. Recurrent models generally factor computation along symbol positions of input and output sequences; however, this sequential process precludes parallelization within training samples. Attention mechanisms have been used to combat this problem, but they are usually used alongside a recurrent model. The Transformer is the first model to rely entirely on an attention mechanism. 

The systems consists of encoder and decoder stacks, each composed of a stack of $N = 6$ identical layers. The attention function maps vector queries, keys, and values, with the output computed as the weighted sum, with weight computed as a function that determines the compatibility of the query with the corresponding key. Learning rate and batch size are used as hyperparameters.

The model has achieved impressive results in machine translation; for example, one model achieved a BLEU score of 28.4 on an English-to-German translation tasks, beating the previous record by 2.0 BLEU points.
\cite{vaswani2017attention}

\begin{figure}[!h]
    \centering
    \includegraphics[scale=0.5]{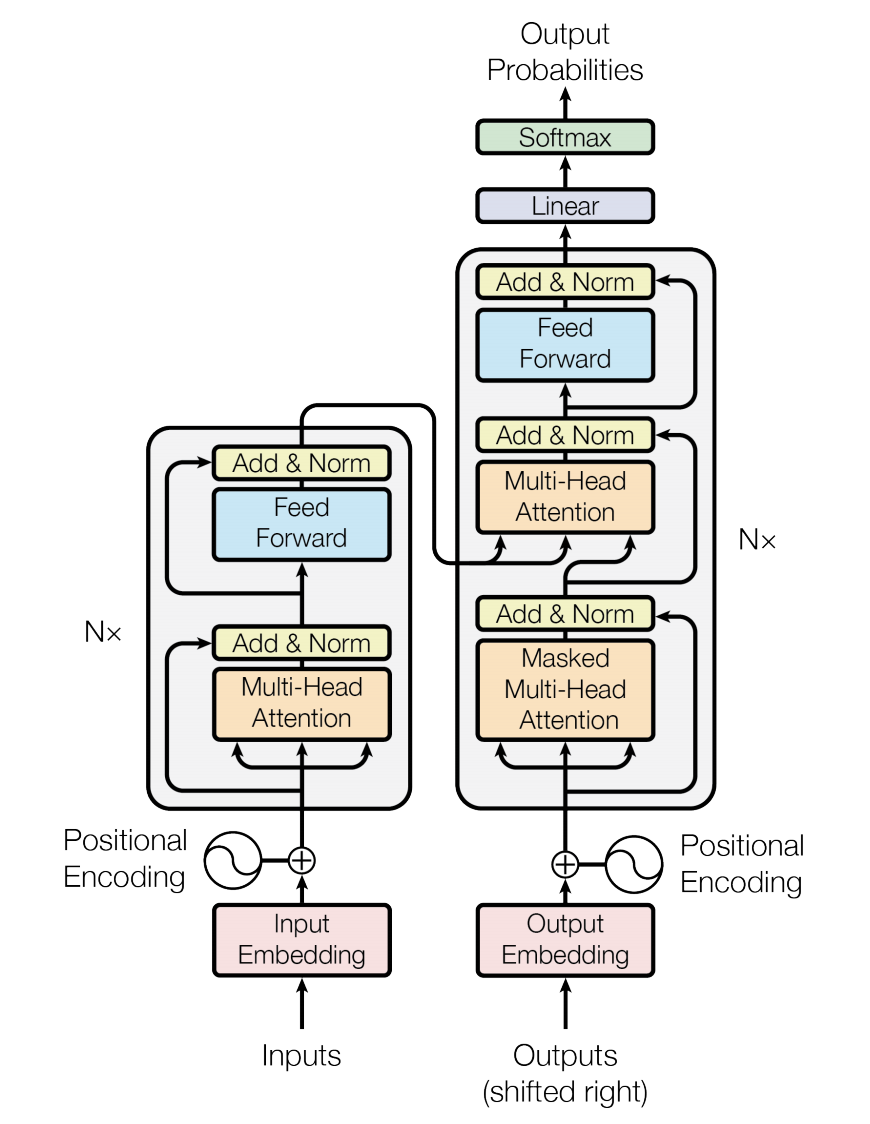}
    \caption{Transformer Model Architecture [2]}
\end{figure}
\subsection{BERT}
The language representation model Bidirectional Encoder Representations from Transformers (BERT) relies on the concept of transfer learning to learn unsupervised from a corpus of unlabeled data. Using a large amount of data allows for BERT to bypass weaknesses of other models, including overfitting and underfitting . Through fine-tuning with a much smaller sample of labeled data, the resultant model can be employed for downstream tasks, such as, question answering, machine reading comprehension, and sentiment analysis. To pre-train, the BERT model masks certain phrases or words from the original input and trains on two prediction tasks: prediction of the masked token words and binary prediction whether the second sentence input belongs after the first in the original text. 
\subsection{ALBERT}
Albert is a more condensed form of BERT, intended to have comparable, or even, superior capabilities as BERT while expending less computational power, and significantly less input time, which makes it a generally more accessible program for users with less time or fancy tech at their disposal. It also allows for a larger model size as it requires fewer resources and can therefore, achieve better performance and accuracy. This has made it popular among corporations, due to the lower cost and training time requirements. The way that Albert manages to work as well as BERT while being significantly faster and less demanding is by having far less parameters, which is reached through  cross-layer parameter sharing, reducing redundancy \cite{lan2020albert}. 
\subsection{XLNet}
One problem with the BERT model is that it neglects dependency between masked positions and also suffers from a pretrain-finetune discrepancy. In order to combat these issues, XLNet was proposed as a generalized autoregressive pretraining method. XLNet enables the learning of bidirectional contexts and overcomes BERT's limitations due to autoregressive formulation, and it outperforms BERT on more than twenty tasks, including tasks in question-answering and sentiment analysis.
Autoregressive language modelling estimates the probability distribution of a sequence of text with an autoregressive model. Since many natural language processing tasks require bidirectional processing, there is a gap in performance between AR pretraining and effective pretraining.
XLNet architecture consists of content-stream attention, which is similar to previously-used self-attention mechanisms, query information, which does not have access to content information, and a two-stream attention model. 
The largest XLNet model consists of the same hyperparameters as BERT and a similar model size; an input sequence length of 512 was always used.
\cite{yang2020xlnet}
\subsection{RoBERTa}
Although self-training models such as XLNet and BERT brought significant performance gains, it is difficult to determine which aspects of the training contributed the most, especially as training is computationally expensive and only limited tuning can be done. A replication of the BERT study that emphasized the effects of training set size and hyperparameter tuning found BERT to be undertrained, and thus RoBERTa is an improved method for training BERT in order to increase performance. Modifications made include training the model longer on larger data sets, changing the masking pattern applied to training data, training on longer sequences, and removing next-sentence prediction. RoBERTa uses the same BERT optimization hyperparameters, with the exception of the peak learning rate and number of warmup steps, which are tuned independently for each setting. The model has proven highly effective, establishing a new state-of-the-art on 4/9 GLUE tasks (MNLI, QNLI, RTE, and STS-B) and matching state-of-the-art results on the RACE and SQuAD datasets. \cite{liu2019roberta}
\subsection{ConvBERT}
While BERT does achieve impressive performance compared to previous models, it suffers large computation cost and memory footprint due to reliance on the global self-attention block. BERT was found to be computationally redundant, since some heads only need to learn local dependencies. The ConvBERT model integrates BERT with a novel mixed-attention design, and experiments demonstrate that ConvBERT significantly outperforms BERT on a variety of tasks; for example, ConvBERT achieved an 86.4 GLUE score, with 1/4 of the training cost of ELECTRAbase. ConvBERT uses batch sizes of 128 and 256 respectively for the small-sized and base-sized model during pre-training, and an input sequence of 128 is used to update the model. \cite{jiang2021convbert}

\subsection{BART}
BART is a denoising autoencoder used for pretraining sequence-to-sequence models. BART is particularly effective when fine tuned for text generation but also works well for comprehension tasks. BART uses a standard Transformer-based neural machine translation architecture, which can be seen as generalizing BERT (due to the bidirectional encoder), GPT (with the left-to-right decoder), and many other pre-training plans.


BART was pretrained by using corrupted documents then optimizing a reconstruction loss. If all information was lost for source, BART would be equivalent to a language model. Although BART does perform similar to the RoBERTa model, it brings new state-of-the-art results on number of  text generation \cite{lewis2019bart}.

\section{Methodology}
\label{sec:approach}


\subsection{BERT Encoding Layer}
 BERT is used as an encoder layer for our ensemble model to achieve tokenized input and utilize the transformer self-attention for a richer encoding than previous embedding models such as Word2Vec \cite{mikolov2013efficient} or Glove \cite{pennington2014glove}. Let $W = (w_1, \dots, w_m)$ be an $m$ word sequence, where $m$ is the max sequence length set as a hyperparameter. We tokenize $W$ using the BERT positional, segment, and sentence embeddings to obtain the tokenized form, $T = (t_1, \dots, t_n)$, of the text sequence. The self-attention blocks in BERT provide for an encoding that accounts for context within the input tokens. Unlike Wang et al \cite{wang2016machine}, we do not use an LSTM preprocessing layer for both the context and the question independently. We find that the bidirectionality of BERT makes for a much more useful encoder than just an LSTM.

 Let $(Q, C)$ denote the question-context pair. The concatenated representation of the question and context will be the input into the BERT-base model. Let $\circ$ represent the concatenation operation. Now, we define the concatenated input sequence as follows:
 \begin{equation*}
     \mathbf{E} = Q \circ C
 \end{equation*}
 The output encoding from BERT can be represented as follows:
 \begin{equation*}
     \mathbf{H} = \text{BERT}(\mathbf{E})
 \end{equation*}
 where $H = (h_1, \dots, h_k)$, with $k$ denoting the output dimension, is the output hidden representation that is computed by encoding the input question-context sequence.
 
\subsection{BiLSTM Layer}
We decided to use a bidirectional LSTM after retrieving the BERT encodings to get more contextual representation from the inputs. Once the BERT encoding of the input sequence is fed into the BiLSTM, the outputs are computed as follows:
\begin{equation*}
    \hat{y}_t = g(W_y[\overrightarrow{a_t}, \overleftarrow{a_t}] + b_y)
\end{equation*}
The LSTM layer efficiently uses past features and future features once inputs have been encoded. While BERT gives contextual encodings with attention among the input sequence, the BiLSTM further brings context without attention. We choose to use the BiLSTM to retain the consistency of bidirectional models. The layer computes two different hidden representations for the input sequence: one for the context from the left and one for context from the right. This mechanism is similar to self-attention in BERT, but doesn't use attention in itself.

  \begin{figure}[t]
    \centering
    \includegraphics[scale=0.5]{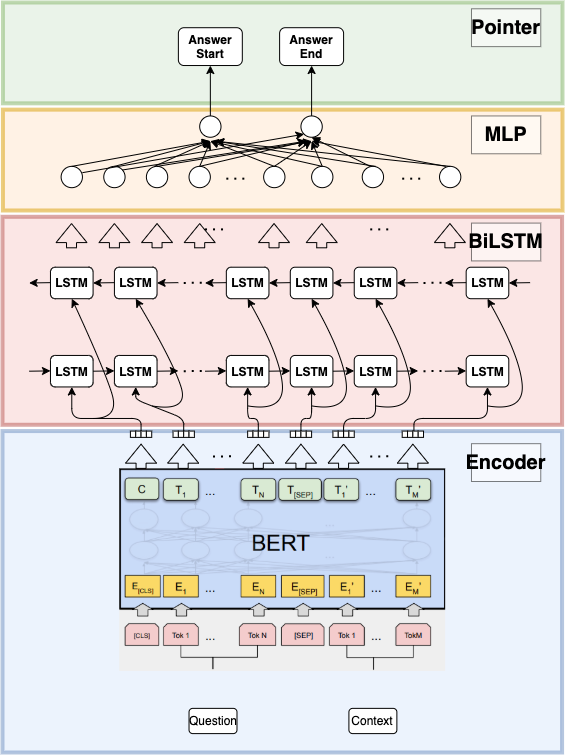}
    \caption{BERT-BiLSTM Model Architecture}
    \label{fig:my_label}
\end{figure}

\subsection{Linear Layer}
To augment the question and context together, we represent the input sequence into the language models as a question-context single packed sequence. The prediction is a start token vector $S \in \mathbb{R}^H$ and an end token vector $E \in \mathbb{R}^H$ that specify the beginning and end of the answer, respectively. Now, the probability of word $w_i$ being the start of the answer span is computed as the dot product between the token representation $T_{w_i}$ (after BiLSTM) and $S$ followed by a softmax over all the words in the sequence:

\begin{equation*}
    P_{w_i} = \frac{e^{S\cdot T_{w_i}}}{\sum_{j} e^{S\cdot T_{w_j}}}
\end{equation*}

The end of the answer span is computed analogously. The score of a candidate span from position $i$ to position $j$ is defined as $S\cdot T_{w_i} + E\cdot T_{w_j}$ and the span with the highest probability where $j\geq i$ is used as a prediction.

\section{Experiments}
\label{sec:experiment}
In this section, we train the various pre-trained language models discussed previously and our proposed ensemble model. We fine-tune our models on the following datasets.

\subsection{Datasets}
\label{sec:datasets}
\subsubsection{\textbf{NewsQA}}
NewsQA is a dataset consisting of 119,633 questions posed by crowdworkers on 12,744 CNN articles. Answers to the questions are contained within spans of text in the articles.

NewsQA differentiates itself from previous datasets with a few characteristics: questions may be unanswerable (having no answer in the corresponding article), and questions require reasoning beyond simple word and context matching. Additionally, there are no candidate answers to choose from, and answers are spans of text of arbitrary length, rather than single words or entities. There are several types of answers in the NewsQA dataset, including people, dates, numeric figures, verb phrases, and more.
\begin{figure}[!h]
    \centering
    \includegraphics[scale=0.4]{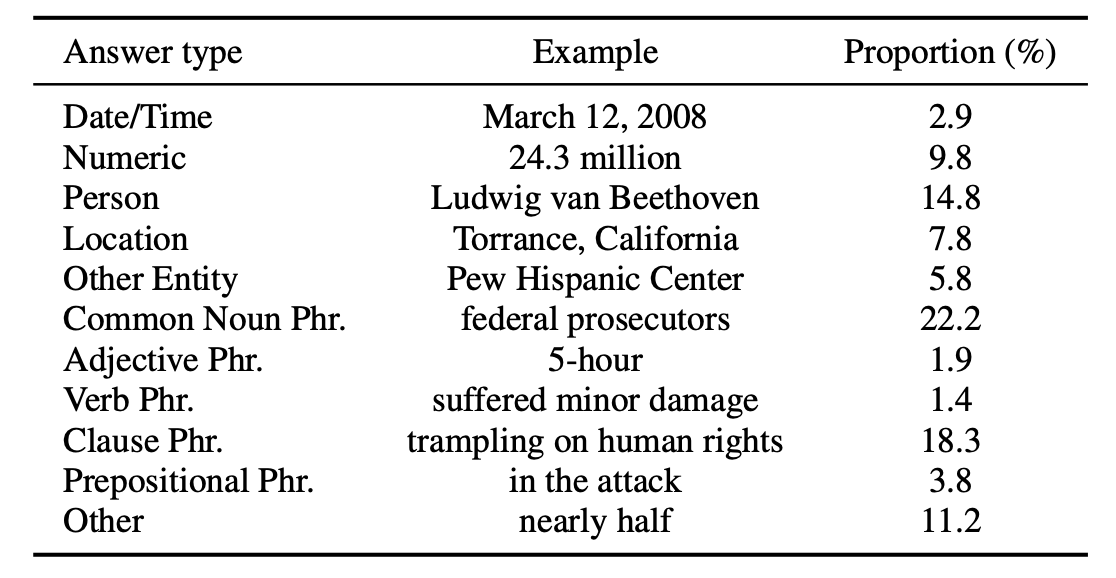}
    \caption{Example of NewsQA question types \cite{trischler2017newsqa}}
\end{figure}

The gap between human performance and model performance on the NewsQA dataset is an astounding 0.198 F1 points, demonstrating a large margin for improvement for future models.
\cite{trischler2017newsqa}

\subsubsection{\textbf{SQuAD 2.0}}
One major flaw in the SQuAD dataset is the focus on questions that are guaranteed to have an answer within the context, making models use context and type-matching heuristics. In order to combat this, the SQuAD 2.0 dataset combines answerable questions from SQuAD with 53,755 unanswerable questions about the same paragraphs. Models that achieve an 0.858 F1 score on SQuAD achieve only a 0.663 F1 score on SQuAD 2.0, creating room for improvement for models.

Unanswerable SQuAD 2.0 questions aim to achieve relevance, in order to prevent simple heuristics like word overlap from being used to determine if a question is answerable, and also the existence of plausible answers in the context, in order to prevent type-matching heuristics from being used to determine if a question is answerable.

SQuAD 2.0 forces models to understand whether an answer to a question exists in a given context in order to encourage the development of reading comprehension models that know what they don't know and have a deeper understanding of language \cite{rajpurkar2018know}.

\begin{center}
\begin{table*}
\centering
 \begin{tabular}{l | c c c c} 
 \hline
   & \textbf{NewsQA} & \textbf{SQuAD} & \textbf{QuAC} & \textbf{CovidQA} \\ [0.5ex] 
 \hline
 XLNet$_{BASE}$   &  53.2   &  64.9 & 30.1 & 44.9 \\
 BERT$_{BASE}$   &  52.1   & 64.7  & 28.6 & 44.8 \\
 RoBERTa$_{BASE}$  &  57.0  & 68.2  & 31.3 & 44.5 \\
 ALBERT$_{BASE}$ &  51.8  & 64.8  & 19.5 & 42.4 \\
 ConvBert$_{BASE}$ &  55.7   &  67.4 & 31.5 & 44.9 \\
 BART$_{BASE}$ &  56.2   &  67.6  & 29.1 & 45.3 \\
 BERT$_{BASE}$-BiLSTM &  52.6   & 65.0 & 28.9 & 45.6 \\
 \hline
\end{tabular}
\vspace{0.2cm}
    \caption{F1-scores for various pre-trained models on NewsQA, SQuAD, QuAC, and CovidQA datasets.}
    \label{tab:my_label}
\end{table*}
\end{center}

\subsubsection{\daffy \textbf{QuAC}}
QuAC is a Question Answering in Context dataset. The dataset consists of two dialogues: one of a student who asks open-ended questions in order to learn as much about a Wikipedia article as possible, and another of a teacher who answers the student's questions using excerpts from the text. QuAC contains about fourteen thousand dialogues, with about a hundred thousand questions in total. The QuAC dataset improves on many previous question-answering datasets; for example, its questions are often unanswerable, open-ended, or only make sense in context. 
\begin{figure}[!h]
    \centering
    \includegraphics[scale=0.4]{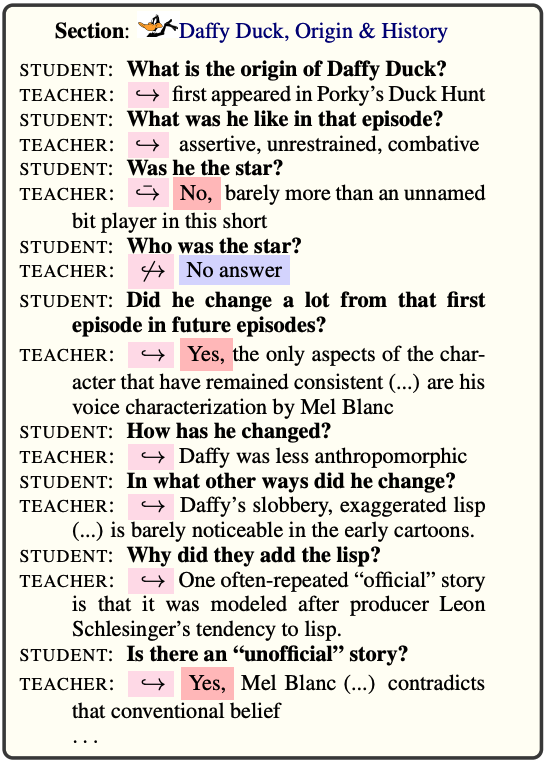}
    \caption{Example of a QuAC dialogue. The context is taken from the section "Origin and History" from the Daffy Duck Wikipedia article. [2]}
\end{figure}

Unlike SQuAD, students do not know the answers to their own questions before posing them, making simple paraphrasing and string-matching less likely. It also contains coreference to previous answers and questions.

The task begins with a student formulating an open-ended question (q) from the information that they have been given. The teacher must answer with a continous span of text given by indices (i,j). In order to create more natural interactions, the teacher must also provide the student with a list of dialogue actions, such as continuation (e.g. follow up), affirmation (e.g. yes or no), and answerability (answerable or no answer)

The QuAC dataset has long answers that average fifteen tokens, compared to three on average for the SQuAD dataset - this is unsurprising, considering that most SQuAD answers are numerical figures or entities, while QuAC questions are typically more open-ended. Although QuAC questions are typically shorter than SQuAD questions, this does not mean the questions are less complex; in the QuAC datasets, students cannot access the context section to paraphrase.\cite{choi2018quac}

\subsubsection{\textbf{CovidQA}}
CovidQA is a Question Answering dataset consisting of 2,019 different question/answer pairs. There were 147 articles most related to COVID-19, pulled from CORD-19, that were annotated by 15 volunteer biomedical experts. Even though the annotators were volunteers it was required for them to have a Master's degree in biomedical sciences. The team of annotator were led by a medical doctor (G.A.R), who assessed the volunteer's credentials and manually verified each of the question /answer pairs produced. The annotations were created in SQuAD style fashion which is where the annotators mark text as answers and formulate corresponding questions.

CovidQA differs from SQuAD being that the answers come from longer texts (6118.5 vs. 153.2), answers are generally longer (13.9 vs. 3.2), and it doesn't contain n-way annotated development nor test sets. RoBERTa-base-architecture and a fine-tuned version of the SQuAD dataset was chosen and used as the baseline model of CovidQA. To train the baseline model the CovidQA annotations were used in a 5-fold cross validation manner. 

\subsection{Preprocessing}
Before implementing our models, we preprocessed the dataset to fit the format required as input to the pre-trained transformers. The datasets were scraped to create (context, question, answer) triplets for each context paragraph and its corresponding questions and answers. The context and question are concatenated for input into the BERT tokenizer, which tokenized the sequence and added the [CLS] token at the beginning of the input sequence and the [SEP] token between the context and the question. For the sake of computational ability, we set the maximum sequence length for all tokenizers to 512. All datasets are tokenized with WordPiece \cite{wu2016googles} and SentencePiece \cite{kudo-richardson-2018-sentencepiece} and uniformed in lower cases.

\subsection{Hyperparameters}
All the models are implemented using PyTorch and the Transformers library. We utilize the base uncased version of each model with maximum sequence length 512. The document stride is set to 128. To optimize the models, we use the Adam optimizer with learning rate 5e-5 and train the models for a total of 3 epochs. We set the batch size for our training and validation sets to 8.
We utilize the base version of each of the pre-trained language models.

The code, model, and datasets are publicly available at for reproducibility.\footnote{\url{https://github.com/Akomand/AEOP_Research_2021}} We performed our experiments on an NVIDIA 2x Quadro RTX 8000 (48G) GPU RAM.

\subsection{Evaluation Metrics}
To evaluate the performance of our fine-tuned models on each of the datasets, we use the F1-score. The F1-score takes into account both the precision and recall of the model. The F1-score is computed for both the classified start token $S$ and the end token $E$ and averaged to get a single F1-score defined in terms of the precision and recall as follows:

\begin{equation*}
    \text{precision} = \frac{TP}{TP + FP}
\end{equation*}

\begin{equation*}
    \text{recall} = \frac{TP}{TP + FN}
\end{equation*}

\begin{equation*}
    \text{F1-score} = 2* \frac{\text{precision}\cdot\text{recall}}{\text{precision}+ \text{recall}}
\end{equation*}

\section{Results}
\label{sec:results}
We evaluate the performance of the base models using the F1-score. We fine-tuned  XLNet$_{BASE}$,  BERT$_{BASE}$,  RoBERTa$_{BASE}$,  ALBERT$_{BASE}$,  ConvBERT$_{BASE}$,  and BART$_{BASE}$ on the NewsQA, SQuAD, QuAC, and CovidQA datasets. The models performed the best on SQuAD 2.0 since the contexts, questions, and answers were straightforward. Further, SQuAD consists of relatively short sequence answers to all questions and the answers are purely extractive. However, each of the fine-tuned models performed quite poorly on the QuAC dataset due to the fact that QuAC consists of quite open-ended questions that require more inference capabilities from the model. The models performed almost as well on the NewsQA dataset as the SQuAD dataset, which demonstrates impressive capabilities of the models since the NewsQA dataset is a challenging machine comprehension dataset that requires reasoning. CovidQA is a small dataset that consists of contexts, questions, and answers of much longer lengths than other datasets. As such, there was not enough training data to help the models generalize better. Since most of the pre-trained language models can only handle a relatively short max sequence length, the model lost a large amount of information when processing contexts, thereby decreasing model performance. We notice that RoBERTa and BART are among the highest perfoming models. RoBERTa is a highly optimized version of BERT that focuses on the masked language modeling pre-training task without the next sentence prediction task. The masked language modelling specifically helps the model perform better than others for the question answering task since the goal is to classify the start and end tokens. Since BART uses a neural machine translation type architecture, BART is able to utilize both the encoder and decoder blocks from transformers for the token classification task. Our model (BERT + BiLSTM) outperforms BERT$_{BASE}$ by at least $1\%$, so an additional layer of bidirectionality helps contextual representations for better performance in question answering. 


\section{Conclusion}
\label{sec:conclusion}
In this paper, we analyze the performance differences between various pre-trained language models fine-tuned on question answering datasets of varying levels of difficulty. Further, we propose an ensemble model and compare its performance to other models. Experimental results show the effectiveness of our methods and shows that RoBERTa and an auxiliary BiLSTM layer both improve model performance in question answering. We see the highest F1-score on RoBERTa and BART model across all datasets. We also observe at least a $1\%$ increase in F1-score over the BERT base model when a BiLSTM layer is added on. Future work includes extending our model to incorporating additional attention mechanisms and potentially utilizing the MatchLSTM architecture to create a better performing ensemble model.

\section{Acknowledgements}
\label{sec:acknowledgements}
This work was supported in part by the Department of Defense and the Army Educational Outreach Program.


\bibliographystyle{IEEEtran}
\bibliography{Bib}

\end{document}